\pdfoutput=1

\documentclass[11pt]{article}

\usepackage[review]{iccv}

\usepackage{times}
\usepackage{latexsym}

\usepackage[T1]{fontenc}
\usepackage{graphics}
\usepackage{graphicx}
\usepackage{lipsum}

\newcommand\blfootnote[1]{%
  \begingroup
  \renewcommand\thefootnote{}\footnote{#1}%
  \addtocounter{footnote}{-1}%
  \endgroup
}

\title{CapsFlow: Optical Flow Estimation with Capsule Networks}

\begin{document}

\maketitle
\begin{abstract}
\blfootnote{Draft completed as part of undergraduate thesis and submitted to ICCV 2019 Neural Architects Workshop. Edited to correct conference name mistake in previous version}
We present a framework to use recently introduced Capsule Networks for solving the problem of Optical Flow, one of the fundamental computer vision tasks. Most of the existing state of the art deep architectures either uses a correlation oepration to match features from them. While correlation layer is sensitive to the choice of hyperparameters and does not put a prior on the underlying structure of the object, spatio temporal features will be limited by the network's receptive field. Also, we as humans look at moving objects as whole, something which cannot be encoded by correlation or spatio temporal features. Capsules, on the other hand, are specialized to model seperate entities and their pose as a continuous matrix. Thus, we show that a simpler linear operation over poses of the objects detected by the capsules in enough to model flow. We show reslts on a small toy dataset where we outperform FlowNetC and PWC-Net models.
\end{abstract}

\section{Introduction}

Optical flow estimation is a fundamental computer vision problem. Given a pair of images, optical flow attempts to find dense motion field, assigning a vector displacement indicating where they moved to in the second image. Estimating the displacement field between the two images requires learning both the finer local details as well as the global structural information to match them at different locations

In recent years, great progress has been achieved in estimating the optical flow using deep learning methods \cite{3, 7, 11, 15}. Although existing approaches have achieved good performance, most methods rely on a multi-stage pipeline and calculation of correlation map or learning spatio-temporal features between image features to aid the matching process. Correlation typically works by calculating correspondences between a set of pixels within a given neighborhood, for a particular search space.

Formally, given two feature maps \(\bf f_{1}, \bf f_{2}\) with \(w,c,h\) being g their width, height and number of channels, correlation layer compares each patch from ${\bf f_{1}}$ with each patch from \(\bf f_{2}\) in a fixed neighbourhood. The size of the path (\textbf{K}) is equal to \(2k + 1\), where \(k\) is the kernel size. Also for reducing complexity, all patches are not matched. A patch centered at \(x_{1}\) in \(f_{1}\) is only correlated with patches in \(f_{2}\) whose center \(x_{2}\) lies in the neighbourhood whose size is ${2d + 1}$ and is centered at \({x_{1}}\), where \({d}\) is the displacement parameter. The complexity of the operation is \({D^{2} × c × w × h}\), and the output is of size \({(w \times h \times D^{2})}\).

Thus, depending upon these hyper-parameters correlation may or may not be able to find an exact match. Similarly, learning spatio-temporal features would be limited
by the receptive field of the filters used. Capsules on other
hand are naturally suited to this task as position and orientation of any entity captured by a particular capsule is represented by a continuous motion vector which can model
various types of complex rigid or non-rigid motion. To this
end, we show the effectiveness of capsules on a toy dataset
and also present first work, to the best of our knowledge, to
extend capsules to dense prediction tasks.

\section{Related Work}

\textbf{Capsule Networks:} Capsule Networks introduced by \cite{14} are an alternative neural net architecture as compared to Convolutional Neural Networks (CNN) that aim for viewpoint equivariance rather than invariance. Capsule Networks do so by using a group of neurons, called capsules, to encode both the presence as well pose of the entity with respect to the viewer. Furthermore, they use a dynamic routing mechanism to transform poses from one capsule layer to the next

Though initially, capsule networks used a single 8- dimensional vector to encode pose of an entity with its norm representing the probability of its presence, it was later modified by \cite{5} to use a 4×4 matrix to represent entity’s pose and a separate scalar to denote presence.  \cite{5} also replaced routing by agreement in \cite{14} with EM based routing. Since their introduction, capsules have been used for binary segmentation \cite{9}, Action Recognition \cite{4}, 3D point reconstruction \cite{19} and many other applications.

\textbf{Optical Flow Methods:}  Classically, the optical flow was estimated via energy minimization methods, particularly after the work of \cite{6}. However, energy minimization methods fail to work for large displacement (\cite{2}) and a series of works (\cite{1, 10, 18}) use descriptor or feature matching in conjunction with energy minimization, carried out in a coarse-to-fine manner, to alleviate the problem. \cite{17} blended the aforementioned work with deep learning, using manually selected convolutional filters to extract features at multiple scales before energy minimization

CNNs have become ubiquitous for high-level vision
problems and optical flow computation is no exception.
\cite{3} introduced FlowNetC, which used a correlation over a set of learned features and refines it using CNN layers in a multi-scale fashion. SpyNet \cite{12} stacks
both the input images together to learn spatio-temporal filters at multiple-scales and learns residual flow to hierarchically refine the output. PWC-Net \cite{15} combines the two
by using the correlation between learned features to get an
estimate of the disparity at multiple scales, followed by hierarchical refinement to produce the final flow. Also, as
getting real-world data for optical flow is a difficult task,
\cite{11, 16, 20} have explored unsupervised methods for the
task and build on FlowNetC’s architecture and use bidirectional warping based loss along with additional smoothness
constraints to achieve satisfactory performance.

Our work is a departure from this basic template as we
try to leverage the better representational capability of capsule networks, owing to the presence of a motion vector, to
implicitly model different kinds of transformations an object may undergo.

\begin{figure*}
\centering
\includegraphics[width=150mm]{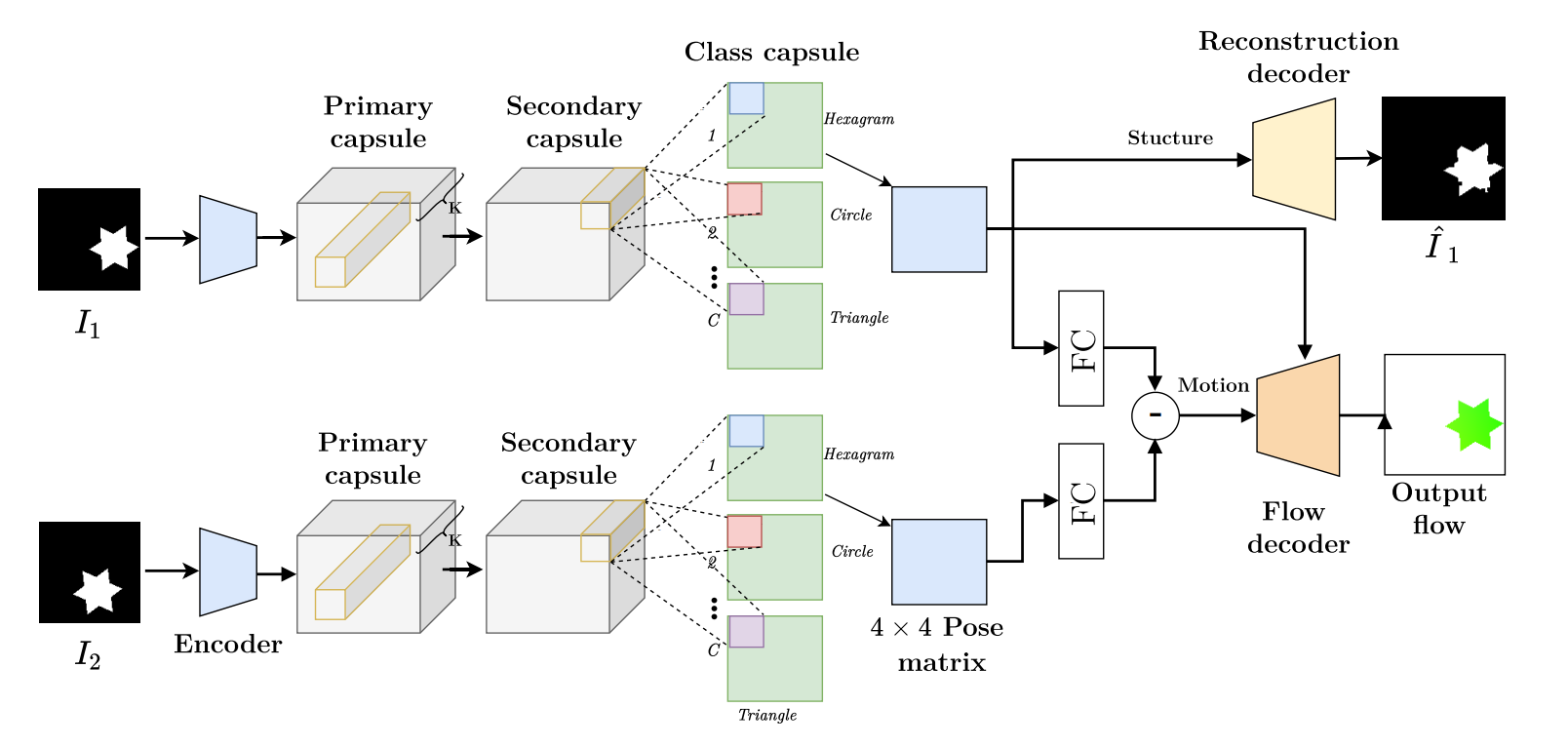}
\caption{First part of the architecture consists of a Siamese network where a pair of input images, Image 1 (\(I_{1}\)) and Image
2 (\(I_{2}\)), are passed through a convolutional encoder. Followed by 3 layers of capsules (including class capsules) to output a
4×4 pose matrix for each of the \(C\) (=5) possible classes. We select the matrices belonging to the classes with the highest
average activation for both images. Then, the two encodings are passed through FC layers to form richer embeddings. We
then subtract the two embeddings to get a motion embedding which is then passed to a convolutional decoder to get the output
flow map. Pose matrices from the first image are also passed as skip connection to flow decoder to refine the flow. The pose matrices are also used to reconstruct back the input image \(I_{1}\) which acts as a regularizer.}
\end{figure*}

\section{Overall Approach}
\textbf{Intuition:} A single neuron in a CNN captures the presence or absence of a feature in the image, on the other hand,
a single capsule captures not only the existence of an entity but also outputs a 4×4 pose matrix which can learn the
spatial relationship (translation, rotation, etc.) of an entity
with respect to the viewer. We utilize this property by calculating the motion of an entity (object) as a transformation
between its pose matrices. This reduces the computationally expensive correlation layer into a single operation. The
intuition for the reformulation is that the pose matrix of a
capsule is capable of learning positional parameters of the
entity it describes, therefore using a suitable transformation
(like subtracting the pose matrix for two images) one could
capture how an entity moved between two images

Capsules are ideally more suited to classification tasks
since each class capsule tries to model a separate entity.
But, real-world scenes and similarly traditional optical flow
datasets do not have a setlist of classes. We thus show the effectiveness of capsule networks in case of class supervision
by using a toy dataset consisting of five different shapes (or
classes) of varying sizes and orientation as shown in Figure
3.

\subsection{Capsule network}
Before we explain our CapFlow architecture, we briefly
explain the main components of a standard Capsule Network:

\textbf{Convolutional encoder:} Capsule networks use few convolutional layers along with ReLU activation to extract basic features from the inputs. These features are then used as
input to primary capsule layers.

\textbf{Capsule Layers:} The output from the convolutional layers is individually passed to a set of capsules, which perform a convolution to output a 4×4 transformation matrix,
M, representing pose of the detected entity and an activation probability, A. The output from the primary capsules is
then routed into secondary capsule layers using a trainable
transformation matrix W between each pair of capsules of
consecutive layers. A pose matrix of capsule \(i\), in layer \(L\),
is multiplied by \(W_{ij}\) to yield \(V_{ij} = M_{i}W_{ij}\) which represents its vote for the \(j^{th}\)
 capsule in layer \(L + 1\). The poses
in \(L + 1\)th layer then uses a non-linear routing mechanism
(explained next) to determine its poses and activation vector. Final capsule layer consists of a single capsule representing the final classification label.

Routing between Capsules: Each of the element in
4×4 pose matrix of the parent capsule belongs to a Gaussian distribution whose parameters are estimated using the
Expectation-Maximization (EM) algorithm. Broadly, Expectation step attempts at calculating the probability that
each of the capsules in layer \(L\) is explained by capsules in
layer \(L + 1\). Maximization step then tries to maximize the
activation of a given capsule in layer \(L + 1\) depending on
inputs from layer \(L\).

\begin{figure}
\centering
\includegraphics[width=80mm]{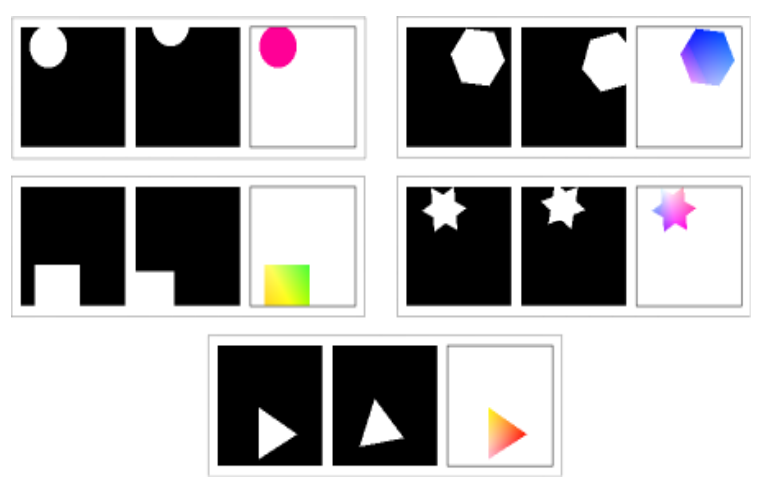}
\caption{Examples of input images and the ground truth
flow used for training and validation.}
\end{figure}

\subsection{CapsFlow network architecture}
The overview of Siamese CapsFlow architecture is
shown in Figure 1. The input to the network is a pair
of 128×128×3 RGB images, I1 and I2. The network
begins with four convolutional layers with a kernel size
of 3×3 and strides of 2 and 1 alternatively, each with
a ReLU non-linearity. The resulting feature map of dimension \textbf{32×32×256} is transformed into a capsule layer
of dimension \textbf{16×16×16×17}, (\textbf{K×H×W×4*4+1}, where
\(K\)=number of capsule types) by applying a 3×3 convolutional operation with stride 2. This is followed by a second
capsule layer with 16 capsule types and a 7×7 receptive field with stride 2.

The second capsule layer is then connected to a final
convolutional capsule layer (class capsule) with C capsule
types, where C is the number of shapes in the dataset. The
resulting output (\textbf{C×6×6×(16+1)}) consists of two components, a pose matrix (\textbf{C×6×6×16}) and an activation vector (\textbf{C×6×6×1}). Note that, we do not perform coordinate
addition to calculate the poses and activations of the final
capsule layer since the penultimate and final capsule layers
are not fully connected therefore there is no loss of spatial
information (\cite{5,4}). Coordinate addition is required when
routing from a 2D capsule layer (\textbf{C x H x W x 17}) to a 1D
capsule (\textbf{C×17}) to preserve spatial information. The activation probability for each shape is calculated by averaging
over the spatial dimensions (\textbf{6×6}) of the shape’s activation
vector. The capsule with the highest activation probability
corresponds to the shape predicted by network.

The output pose of the class capsule layer (C×6×6×16)
is then masked to obtain the pose matrix corresponding to
the ground truth shape. During training, we use the knowledge of ground truth to choose the correct class while during
inference the class is chosen based on the highest mean activation vector. The masked output (6×6×16) is passed to
a fully connected layer with RELU non-linearity to obtaine structural embedding of shape (8×8×16). The structure
embeddings for both \(I1\) and \(I2\) are passed to a shared convolutional layer and then subtracted to calculate the motion
embedding, which encodes the motion from \(I_{1}\) to \(I_{2}\).

To help the network learn the correct pose parameters,
we pass the structure embedding to a four layer transposed
convolutional decoder to reconstruct the input image, \(I_{1}\).
This is necessary since training without a reconstruction decoder results in the network learning a sub-par structure embedding, which in turn affects the motion embedding. To
construct the final flow map, both the motion and structure
embedding are passed to a decoder consisting of four transposed convolutional layers each with stride 2 and kernel
size of 4×4. The output of the decoder (64×64×2) is extrapolated to obtain the predicted optical flow map of shape
128×128×2.

\subsection{Objective Function}

\begin{figure}
\centering
\includegraphics[width=80mm]{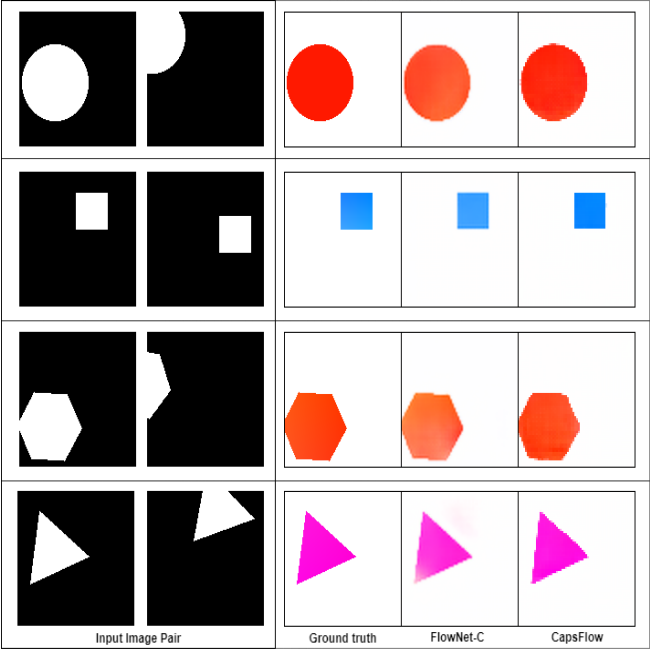}
\caption{Few examples indicating FlowNetC underestimating flow while output of our proposed model is much closer to ground truth. (Best seen in color and zoomed in.)}
\end{figure}

The objective function is made of three losses: a) Mean
squared reconstruction loss to learn the correct pose embedding, b) Activation loss to learn the correct target class, and
c) End-Point-Error (EPE) loss between ground truth flow
map and predicted flow map. To calculate reconstruction
loss, we compute mean square error between \(I_{1}\) and reconstructed image \(\hat I_{1}\).

    \[{L_{mse}} =  || I_{1} - {\hat I_{1}} ||^{2}\]

Empirically, we found that reconstruction loss acts as a regularizer and helps ensure that the pose matrix learns all the
relevant features. In absence of reconstruction loss, the network does not converge, probably due to the reason that
there is not enough supervision to learn correct pose matrices (flow depends only on their difference and not actual
values). We use margin loss to train the activation vector
and is calculated as,

    \[{L_{m}} =  \sum_{i \neq t} max[0, m-(mean(a_{t}) - mean(a_{i})]]^{2}\]

where \(a_{t}\) is the activation of the target shape class and ai
is
the activation corresponding to capsule \(i\). We take the mean
of the activation before computing the loss. The capsule
with the highest activation is used at test stage to determine
the poses to be subtracted to estimate flow. The margin \(m\) is
set to 0.95 during training. EPE loss is calculated as

    \[{L_{epe}} = || flow_{gt} - flow_{pred} ||^{2}\]

Thus, CapsFlow network is trained using the following objective function

    \[{L} = L_{epe} + \lambda_{1}L_{margin} + \lambda_{2}L_{mse}\]

\section{Experiments}

\begin{table}
\begin{center}
\begin{tabular}{ c | c c c c c} 
 \hline
 \multicolumn{6}{c}{Encoder} \\
 \hline
 Conv Stride & 3x3 & 3x3 & 3x3 & 3x3 & \\
 No.of filters & 2 & 1 & 2 & 1 & \\
 BatchNorm & Yes & Yes & Yes & Yes & \\
 \hline
 \multicolumn{6}{c}{Decoder} \\
 \hline
 Transpose Conv & 4x4 & 4x4 & 4x4 & - & - \\
 Convolution & - & - & - & 1x1 & 1x1 \\
 Stride & 2 & 2 & 2 & 1 & 1 \\
 No.of filters & 64 & 64 & 32 & 16 & 2 \\
 \hline
\multicolumn{6}{c}{Reconstruction Decoder} \\
\hline
Transpose Conv & 4x4 & 4x4 & 4x4 & 4x4 & - \\
 Convolution & - & - & - & - & 1x1 \\
 Stride & 2 & 2 & 2 & 1 & 1 \\
 No.of filters & 32 & 64 & 16 & 8 & 1 \\
 \hline
\end{tabular}
\end{center}
\caption{CapsFlow network architecture}
\end{table}
To check the effectiveness of our approach we prepare
a toy dataset consisting of simple shapes as shown in Figure 3 and compare our method to FlowNetC \cite{3} as well as
PWC-Net \cite{15}. We also explore ways to extend our method
to real-world images by experimenting on Flying Chairs
dataset

\subsection{Training}
The data is generated by sampling a random shape type
(square, hexagon, triangle, circle, and hexagram), and a random \((x, y)\) location between 0 to 128. The shape is then
randomly rotated and translated to obtain the corresponding second image. The data is generated on-fly to prevent
overfitting. We have a fixed set of 2500 images used for
testing. The architecture details of encoder and decoders
used is provided in Table 1.

The reconstruction and flow decoder can either be
trained simultaneously or sequentially, starting with first
with reconstruction and then flow. The values of \(\lambda_{1}\) and \(\lambda_{2}\) in equation 4 are chosen to be 0.05 and 2.5 respectively. All models were trained on an NVIDIA GTX 1080Ti GPU. CapsFlow was trained for a total of 30K iterations with a mini-batch size of 64. For sequential case, the reconstruction decoder was first pretrained for 10k iterations. The number of EM iterations during routing was set to 3 as in \cite{5}. We used Adam optimizer \cite{8} with a fixed learning rate of 0.001.

\subsection{Results}

\begin{table}
\begin{center}
\begin{tabular}{ c c c c} 
 \hline
 Model & EPE & Params(in M) & Time(ms) \\
 \hline
 CapsFlow (T) & 0.48 &  1.62 & 13.39\\
 CapsFlow (S) & 0.39 & 1.62 & 13.39\\
 FlowNetC & 0.50 & 39.17 & 3.1\\
 PWC-Net & 0.47 & 8.75 & 12.1\\
 \hline
\end{tabular}
\end{center}
\caption{ Quantitative comparison on single shape dataset: T \& S refers to simultaneous and sequential training respectively, for more information refer to Section 4.1. The average time is calculated for an input pair of resolution 128x128. The best result is highlighted in red color, while the second-best in blue color.}
\end{table}

We compare our results with FlowNetC and PWC-Net.
Both of these models were trained using End Point Error
for 90k iterations with a batch size of 16, using Adam Optimizer with learning rate 0.001, similar to proposed CapsFlow. The end-point error comparison for both CapsFlow
models is shown in Table 2. Both variants of CapsFlow form better than both FlowNetC and PWC-Net while having
8× fewer parameters. But, despite lower number of parameters, capsules take significantly more time primarily due to
iterative EM routing procedure. But as capsule networks develop and see wider adoption there is a high likelihood that
hardware acceleration and better routing procedures will be
able to reduce the time overhead.

On visualizing flow outputs for CapsFlow and FlownetC,
we notice that latter underestimates flow magnitude for a
large number of samples as shown in Figure 3. To confirm
this, we test both models on another validation set having
an average flow magnitude 1.5 times more that of the training set.  FlownetC saw a 3x jump in the EPE from 0.50 to
1.51 on this new set while CapsFlow EPE only increased
from 0.39 to 0.81. This highlights a higher generalization
capacity of CapsFlow. We hypothesize the reason for this
observation to be the fact that while capsules model flow
as a continuous motion matrix, correlation is sensitive to
the choice of displacement and neighborhood size and thus
may not be able to model the large magnitude of flows as
well, particularly if they were not seen during training. We also try to see what pose matrices and the difference
matrix model by scaling their magnitude uniformly in the
range [-1,1] in with increments of 0.5. On changing the
magnitude of the difference we see the flow changing from
one extreme in the flow spectrum to total opposite but no
change in overall structure (Figure 4). On the other hand,
the changing magnitude of skip connection in the range
[0.25, 2.5] in increments of 0.25; changes the scale of the
shape as shown in Figure 5. This shows that flow and pose 
are essentially getting disentangled with the flow coming from the difference of poses while structure comes from Image 1’s pose matrix.

\textbf{Extending to multiple shapes:} We generate instances
of our MultiShape dataset by sampling two different shapes
on-fly for each image. The flow for both shapes is combined to produce the ground truth flow map. On average, the
bounding box for each shape is of size 50×50, and the center of both shapes is bounded inside a region of size 90×90.
Therefore the bounding boxes for both shapes on average
have a 45\% overlap.

While training CapsFlow on multiple shapes we mask
one class capsule at a time and use its pose matrix to construct the corresponding flow map. The final output of the
network is obtained by superimposing both flow maps (in
regions where both maps intersect, we consider the flow of
only one shape based on the same priority that was used
for creating the training set). The training parameters and
losses for this experiment are the same as those for a single shape. We achieve a marginally better EPE of 1.78 as
compared to FlowNetC’s, 1.9. For test images, we pick the
two most active class capsules and use their pose matrices
to construct the individual flow map (which are later used
to find the final flow map). One advantage of CapsFlow is
that unlike other methods we can obtain the individual flow
maps for each shape in the image. Since capsules enforce
a strong constraint on shapes of each object, even in cases
where one of the shapes is highly occluded, CapsFlow can
extrapolate the whole object using its prior about the shape
of the object as shown in Figure 6.

\begin{figure}
\centering
\includegraphics[width=70mm]{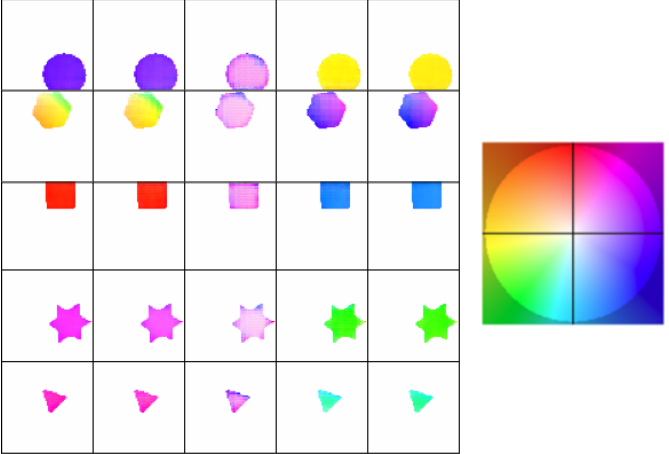}
\caption{\textit{Left}: Results from interpolating flow. We scale
the magnitude of difference matrix between the poses of
given images while keeping the skip connection from I1
same and notice that difference matrix only changes the
flow without any change in structure. To be precise, scaling
magnitude seems to be equivalent to moving from one end
of flow spectrum to diametrically opposite. \textit{Right}: Optical
flow color coding.
}
\end{figure}

\subsection{Unsupervised Training}
We also attempt to test the effectiveness of the proposed
model in an unsupervised setting. To train CapsFlow in
an unsupervised setting, we average the outputs of all the
capsules in the final layer, rather than masking them. We train CapsFlow using the same set of losses used in UnFlow [\cite{11}], the losses are a) occlusion-aware photometric
loss between the input images and their flow warped counterparts, b) second-order smoothness constraint to encourage co-linearity of neighboring flows, and c) forward-backward consistency penalty. Our model performs comparably
to UnFlow method in shapes dataset with an EPE of 2.05 as
compared to UnFlow method EPE of 1.8. But, despite good
EPE score, we notice that in the absence of masking, only
1 out of the 5 classes ever remain active and comparable results can be achieved if other capsules are not considered.
The reason for this is elaborated in the next section

\begin{figure}
\centering
\includegraphics[width=70mm]{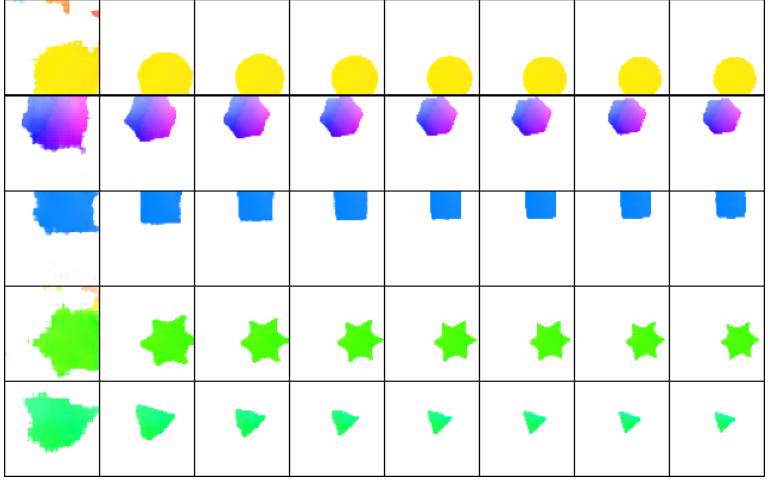}
\caption{We change the magnitude of the pose matrix from
\(I_{1}\) without changing the difference and notice only the output shape of the flow changes without any change to flow
values.
}
\end{figure}

\section{Drawbacks and Future Work}

\begin{figure*}
\centering
\includegraphics[width=150mm]{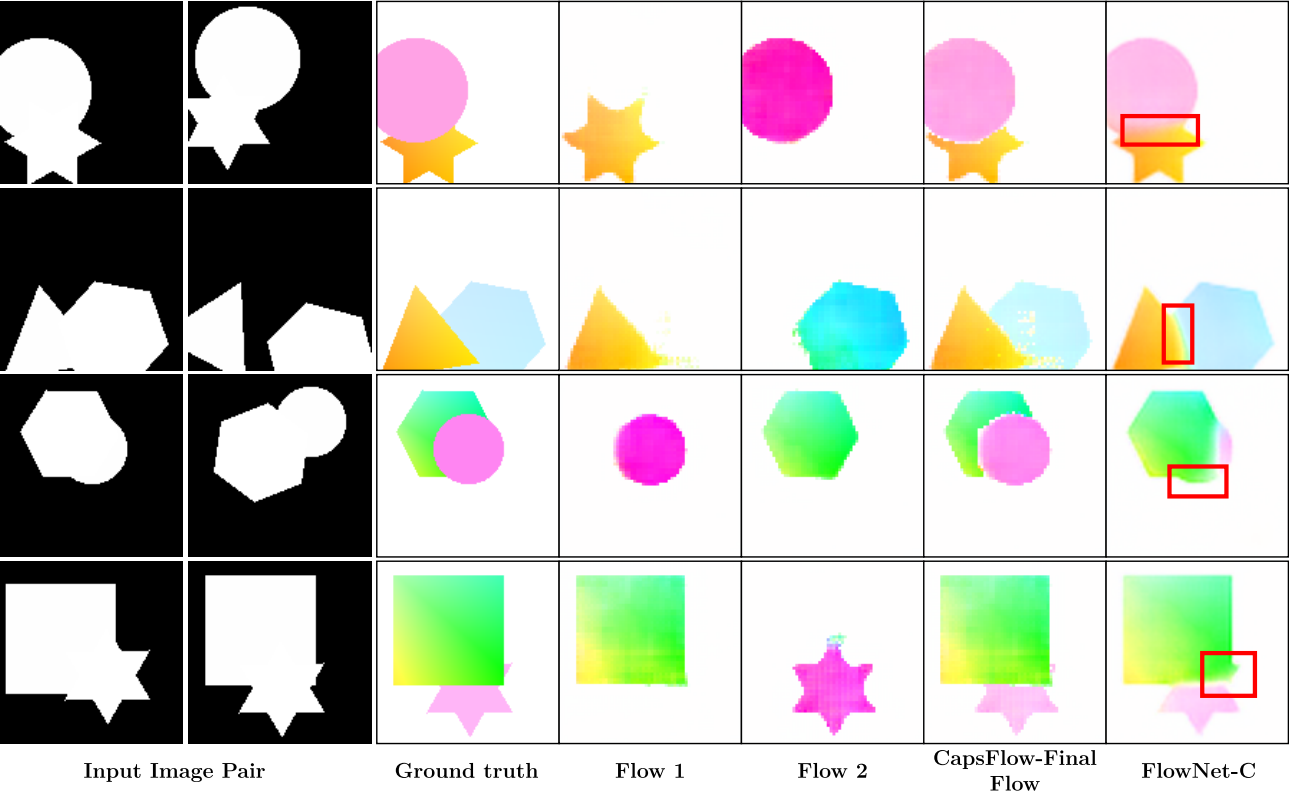}
\caption{CapsFlow can successfully model individual flows even in case of high overlap. Also, at motion boundaries and
overlapping regions, FlowNetC expectedly in absence of any prior on shape boundaries, assigns flow belonging to one shape
to another (highlighted in the red box)}
\end{figure*}

\begin{figure*}
\centering
\includegraphics[width=150mm]{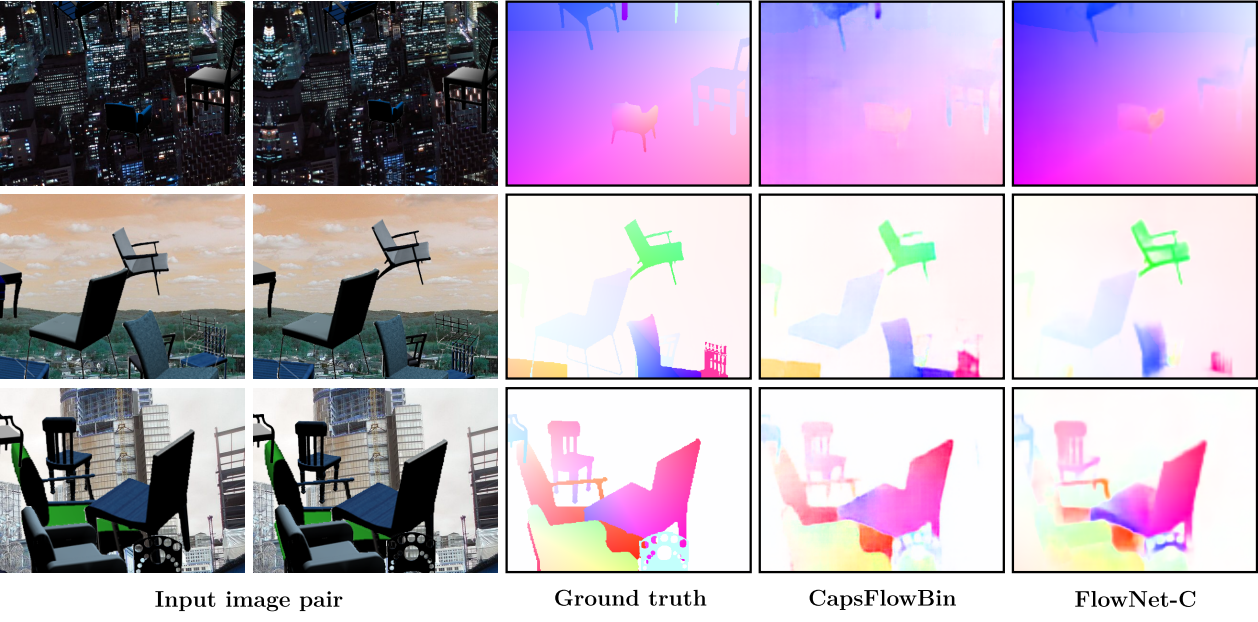}
\caption{Qualitative results on FlyingChairs dataset. Our CapsFlowBin (CapsFlow with binning) method obtained EPE of
3.6, while FlowNetC achieves EPE of 2.0.}
\end{figure*}

Though our proposed model gave promising results on
a toy dataset, capsule network still cannot be naively used
on real world samples. Capsules tend to perform poorly in
tasks where latent capsules do not represent distinct classes
[13]. The sub-par performance is due to capsules learning
parameters using different paradigms. Rather than learning all parameters using EM-routing (generatively), capsules learn transformation matrices in a discriminative fashion through back-propagation in the absence of which we
may get degenerate solutions, such as when all transformation matrices collapsing to zero 

Without concrete class capsules, transformation matrices
have no incentive or gradient to differ from each other. In
this case, the probability that a particular subset of lower
level capsules are transformed into similar clusters in the
pose space of two higher-level capsules \(j\) and \(k\) increases.
This results in a breakdown in the part-to-whole relationship as both \(j\) and \(k\) now model very similar entities, which
in turn results in lower level capsules that route to \(j\) and \(k\)
to model similar entities too [\cite{13}]. Therefore unmasked latent capsule networks require exponentially more capsules
than their masked counterpart. This is a major shortcoming
as such a requirement is very restrictive since the majority of dense prediction tasks have no (definite or explicit)
classes/entities, e.g. Depth estimation, Super-resolution,
3D segmentation. Another drawback of capsule networks
is that, unless explicitly trained, each class capsule models only single instance of that entity, thus will fail in cases
there are multiple instances of same entity in a scene. Further capsules also cannot model global motion though that is
not a major issue as there are other robust techniques available to estimate camera motion.

One work-around for dense optical flow that we tried was
to spatially bin the ground truth and treat those bins as class
capsules. The input to these class capsules was obtained by
subtracting poses from secondary capsule layer itself. This
technique gave good results during training but fails during
testing due to high mis-classification rate (See Figure 7).

But despite these drawbacks capsules seem to be a
promising direction for future research and can address the
drawbacks faced by current CNN based deep methods in
tracking and optical flow computation.

\section{Conclusion}

From the experiments discussed above, we conclude that
Capsule Networks, despite various weaknesses and still being in infancy, offer a great way for modeling motion and
calculating optical flow. Though they outperform the current state of the art on a toy dataset for both single and multiple shapes scenarios and also demonstrate higher generalizability, they require more research to make them adapt to cases where a comprehensive list of entities may not be
available or if there are multiple instances of the same entity
is present in the scene. But despite that, we believe Capsule
Network, due to their greater representational capacity and
implicit encoding of spatial relationships between objects,
are an exciting direction for further research on motion estimation and optical flow. We think that with further improvements in Capsule Networks, our framework can be adapted
to obtain better results on real-world images.

\bibliography{custom}
\bibliographystyle{acl_natbib}

\end{document}